\documentclass[10pt,letterpaper]{article}
\usepackage[a4paper, total={6in, 8in}]{geometry}
\usepackage{amsmath}
\usepackage{amssymb}
\usepackage{amsmath,bm}
\usepackage{amsthm}
\usepackage{url}
\usepackage{graphicx}
\usepackage{graphics}
\usepackage{calligra}
\usepackage{calrsfs}
\usepackage{graphicx}
\usepackage{algorithm}
\usepackage{algorithmic}
\usepackage{epstopdf}
\usepackage{comment}
\usepackage[]{todonotes} 
\usepackage{float}
\usepackage[]{siunitx}
\usepackage{cite}
\usepackage{booktabs}
\usepackage[]{siunitx}

\makeatletter
\def\@@number#1{#1}
\makeatother
\DeclareMathAlphabet\mathbfcal{OMS}{cmsy}{b}{n}

\newcommand{\argmin}{\mathrm{arg}\min}

\newcommand{\PartDeriv}[2]{\frac{\partial{#1}}{\partial{#2}}}
\newcommand{\vect}[1]{\bm{#1}}

\newcommand{\transpose}[1]{{#1}^\intercal}
\newcommand{\derivsym}[1]{\,d{#1}}
\newcommand{\yoavcomment}[1]{}
\renewcommand{\yoavcomment}[1]{#1} 

\newcommand{\OpDistance}{\bm{W}}

\newcommand{\MaskSun}{\mathbfcal{M}}
\newcommand{\Laplacian}{\mathbfcal{L}}
\newcommand{\OpDiag}[1]{\mathbb{D}\left\{#1\right\}}
\newcommand{\DistSet}{\mathcal{C}}

\definecolor{Gray}{gray}{0.9}
\newcommand\authnote[1]{\textsuperscript{\normalfont#1}}
\newcommand\affilnote[1]{\textsuperscript{\normalfont#1}}
\providecommand\textsuperscript[1]{$^{#1}$}

\begin{document}
\title{In-situ multi-scattering tomography}
\date{December 2, 2015}
\author{Vadim~Holodovsky,\authnote{1*} Yoav~Y.~Schechner,\authnote{1} Anat~Levin,\authnote{2}\\
Aviad~Levis,\authnote{1} Amit~Aides,\authnote{1}\\
 \\
\affilnote{1}Dept of Electrical Engineering,\\
 Technion - Israel Institute of Technology,\\
Haifa 32000, Israel\\
\\
\affilnote{2}Dept of Mathematics and Computer Science,\\
 Weizmann - Institute of Science,\\
Rehovot, 76100 Israel\\
{\tt\small \{vholod@tx.technion.ac.il\}}
}
\maketitle

\abstract
To recover the three dimensional (3D) volumetric distribution of matter in an object, images of the object are captured from multiple directions and locations. Using these images tomographic computations extract the distribution. In highly scattering media and constrained, natural irradiance, tomography must explicitly account for off-axis scattering. Furthermore, the tomographic model and recovery must function when imaging is done in-situ, as occurs in medical imaging and ground-based atmospheric sensing. We formulate tomography that handles arbitrary orders of scattering, using a monte-carlo model. Moreover, the model is highly parallelizable in our formulation. This enables large scale rendering and recovery of volumetric scenes having a large number of variables. We solve stability and conditioning problems that stem from radiative transfer (RT) modeling in-situ.


\vspace{-0.2cm}
\section{Introduction}
\label{sec:intro}
\vspace{-0.1cm}

Recovering scenes via participating media~\cite{fattal2008single,fuchs2008combining,namer,narasimhan2005structured,schechnerSpace,treibitz} often focus on observing background objects~\cite{tian2009seeing,morris2005dynamic,zhu2011stabilizing}. However, it also important to recover the medium itself, as done in remote sensing of the atmosphere.
Recent works seek volumetric recovery of a three dimensional (3D) heterogeneous scattering media, focusing on the atmosphere. Being very large, recovery of the atmosphere generally requires passive imaging,
using the steady, uniform and collimated Sun as the radiation source.  The data is images acquired from multiple directions~\cite{veikhermanclouds}, which sample the scene's light-field.

Based on multi-view image data, computational tomography (CT) yields 3D volumetric recovery in many domains~\cite{atcheson2008time,ma2014transparent,trifonov2006tomographic,wetzstein2011refractive}, including biomedical imaging. However, in most CT models, as in X-ray, direct-transmission~\cite{modregger2014multiple} forms the signal, while small-angle scattering has been considered to be a perturbation. In contrast, in a medium as the atmosphere, the source (unidirectional sun) and detector (wide angle camera) are generally not aligned: {\em scattering including high orders is the signal.}\footnote{In a {\em single scattering} regime, each light ray changes direction at most once due to scattering. In a multiple scattering regime, light can change direction by scattering in multiple events (orders).}
For this reason, there is a need to formulate tomography based explicitly on multiple scattering. This work advances this formulation.

Ref.~\cite{aides2013}  assumes that the signal is mostly a result of single-scattering, thus deriving single-scattering tomography. However in many real situations high order scattering is not negligible. Our current work generalizes~\cite{aides2013} to multiple scattering. On the other hand, Refs.~\cite{boas2001imaging,gibson2005recent} focus on a diffusion limit, equivalent to infinite scattering orders. However, many scenes of interest do not comply with the diffusion limit, having regions of low-order scattering. The tomography work we now describe handles any order of scattering events, due to use of a monte-carlo (MC) model. The approach of~\cite{levis2015} performs tomography, where radiative transfer is based on a discrete ordinate spherical harmonic method. While MC can realize scattering events in any location and direction, discrete ordinate methods are by definition constrained to discretized or band-limited propagation. The medium in~\cite{levis2015} is captured from afar: there are no scattering events near the camera.

This paper derives multi-scattering 3D tomography, where cameras can be arbitrarily close to the medium, and in fact can be in-situ, as the system in~\cite{veikhermanclouds}. As we explain, such a setup imposes instabilities on the image formation forward model, which can strongly affect recovery. Moreover, a forward model involving multiple scattering is computationally complex. Attempting an inverse-problem generally magnifies computational complexity, jeopardizing its realistic prospects. This work addresses all these issues. First, we propose a way to stabilize the forward model, including in-situ image rendering, while being computationally efficient. Efficiency is achieved using a forward-MC principle, which parallelizes calculations to multi-view cameras, for each photon packet. This gain is in addition to the inherent parallelizable nature of MC, where each photon packet can traverse the medium independent of other packets. Second, in the inverse problem, we use efficient optimization using surrogate functions~\cite{levis2015}, while solving an ill-conditioned formulation that arises from the in-situ setup.

\section{Theoretical background}
\label{sec:theor-backgr}

In this section we describe the basic building blocks of radiative transfer through a non-emitting medium.  Using these building blocks, two common Monte-Carlo (MC) methods
are described, each having a specific advantage and disadvantage, complementing to the other method, in the context of multiview in-situ imaging. Consequently, in Sec.~\ref{sec:PFM}, we derive a new MC method which better addresses the setup.\\

\noindent {\bf Extinction}: Radiance is a flow of photons. Light propagation through the atmosphere is affected by interaction with air molecules and aerosols (airborne particles). Atmospheric constituents have an {\em extinction cross section} for interaction with each individual photon. Per unit volume, the {\em extinction coefficient} due to aerosols is $\beta^{\rm aerosol} = \sigma^{\rm aerosol} n$. Here $\sigma^{\rm aerosol}$ denotes aerosol extinction cross section and  $n$ denotes particle density. The total extinction is a sum of the aerosol and molecular contributions, $\beta= \beta^{\rm aerosol}+\beta^{\rm air}$, where $\beta^{\rm air}$ is modeled as a function of altitude and wavelength $\lambda$~\cite{aides2013}. The {\em optical depth} along a photon path $S$ is
\begin{align}
  \tau= \int_{S} d\tau=\int_{S} (\beta^{\rm aerosol}+\beta^{\rm air}) dl =\int_{S}
  (\sigma^{\rm aerosol} n +\beta^{\rm air}) dl = \tau^{\rm air}+ \int_{S}
  \sigma^{\rm aerosol} n dl \;,
\label{eq:tau}
\end{align}
where $\tau^{\rm air}=\int \beta^{\rm air} dl$.  The fraction of radiation power that gets transmitted through the atmosphere is the {\em transmittance} $t$, which exponentially decays with the optical depth (Beer-Lambert law):
\begin{align}
  t=\exp(-\tau) \;.
\label{eq:beer-lambert}
\end{align}
\begin{figure*}[t!]
  \begin{center}
    \yoavcomment{\includegraphics[width=\linewidth]{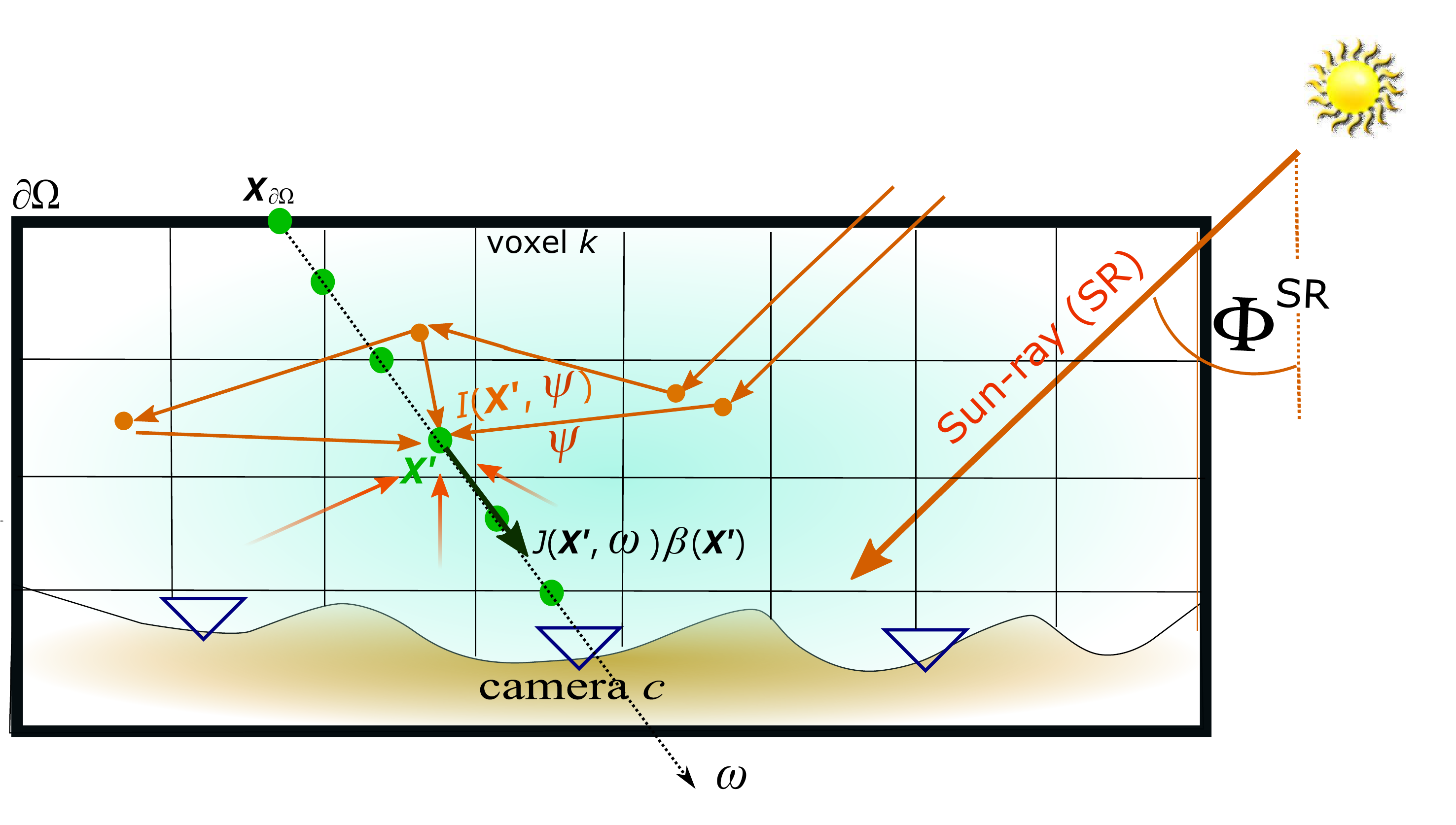}}
  \end{center}
  \caption{\small Integral (lightfield) imaging through a volumetric
    distribution in the atmosphere, using ground-based cameras.}
  \label{fig:setup}
\end{figure*}

\noindent {\bf Scattering}: Suppose that a photon interacts with a single particle. The unitless {\em single scattering albedo} $\varpi$ of the particle, determines a probability for scattering. The aerosol single scattering albedo is $\varpi^{\rm aerosol}$.  The {\em scattering coefficient} due to aerosols in the volume is $\alpha^{\rm aerosol}= \varpi^{\rm aerosol}\beta^{\rm aerosol} =\varpi^{\rm aerosol} \sigma^{\rm aerosol} n $. For non-isotropic scattering, an angular function defines the probability of photons to scatter into each direction. Let ${\bm \omega},{\bm \psi}\in \mathbb{S}^2$ (unit sphere) represent photon or ray directions. The fraction of energy scattered from direction ${\bm \psi}$ towards direction ${\bm \omega}$ is determined by a {\em phase function} $P({\bm \omega}\cdot{\bm \psi})$. The phase function is normalized: its integral over all solid angles is unity, and is often approximated by a parametric expression. Specifically, the Henyey-Greenstein function, parameterized by an \emph{anisotropy parameter} \mbox{ $-1\ge g \ge 1$}, can approximate aerosol scattering~\cite{aides2013}

\begin{equation}
\label{eq:Phg}
P_{\rm HG}({\bm \omega}\cdot{\bm \psi})=\frac{1}{4\pi} \frac{1-g^{2}}{\big[ 1+g^{2}-2 g ({\bm \omega}\cdot{\bm \psi}) \big]^{\frac{3}{2}}}.
\end{equation}
Scattering by air molecules follows the {\em Rayleigh} model
\begin{equation}
\label{eq:Pray}
P_{\rm Ray}({\bm \omega}\cdot{\bm \psi})=\frac{3}{16\pi}\Big[ 1+ ({\bm \omega}\cdot{\bm \psi})^{2} \big].
\end{equation}
In the visible range, air single scattering albedo is $\varpi^{\rm air}\simeq1$ and emission is negligible. For simplicity, wavelength dependency is omitted.
\\

\noindent {\bf Radiative Transfer Equation}:
Denote the volumetric extinction field at position ${\bf X}$ by $\beta({\bf X})$. 
The radiative transfer equation (RTE) describes the flow of radiance $I({\bf X},{\bm \omega})$ at ${\bf X}$, through the medium~\cite{chandrasekhar1960}
\begin{equation}
\label{eq:JbetaRTE}
\nabla_{{\bm \omega}}I({\bf X},{\bm \omega})=-\beta({\bf X})I({\bf X},{\bm \omega})+\beta({\bf X})J({\bf X},{\bm \omega}).
\end{equation}
Here $J({\bf X},{\bm \omega})$ is the {\em in scattering}~\cite{chandrasekhar1960} volumetric field
\begin{equation}
\label{eq:JEq}
J({\bf X},{\bm \omega})=\varpi \int_{4\pi}P({\bm \omega},{\bm \psi})I({\bf X},{\bm \psi})d{\bm \psi}.
\end{equation}
Denote the medium boundary by $\partial\Omega$ and the boundary radiance as $I_{\partial\Omega}$. Let ${\bf X}_{\partial\Omega}$ be the intersection point of the boundary with a ray in direction ${\bm \omega}$.
Integrating Eq.~(\ref{eq:JbetaRTE}) along direction ${\bm \omega}$ defines the integral form of the RTE (Fig.~\ref{fig:setup})
\begin{equation}
\label{eq:IRTE}
I({\bf X},{\bm \omega})=I_{\rm \partial \Omega}{\rm exp}\Big[-\int_{{\bf X}}^{{\bf X}_{\partial\Omega}}\beta(r)dr\Big]+\int_{{\bf X}}^{{\bf X}_{\partial\Omega}}J({\bf X}',{\bm \omega})\beta({\bf X}'){\rm exp}\Big[-\int_{{\bf X}}^{{\bf X}'}\beta(r)dr\Big]d{\bf X}'.
\end{equation}
MC is a popular numerical approach to solve Eqs.~(\ref{eq:JbetaRTE},\ref{eq:JEq},\ref{eq:IRTE}).

\subsection{Monte Carlo Photon Tracking}
\label{subsec:MCPT}

MC methods trace propagated photons, given the source radiance and $\varpi,P({\bm \omega},{\bm \psi}),\beta$. The propagation realizes Eq.~(\ref{eq:IRTE}).  The result is an estimate of the radiance around $({\bf X},{\bm \omega})$, across the domain. In our case study,
the light source (sun) is effectively located at infinity and light is captured by cameras.
A modeled camera $c$ has center of projection at location ${\bf X}_{c}$. Each pixel $p$ collects radiation flowing from a narrow cone around direction ${\bm \omega}_{p}$, yielding
a raw image $i_c({\bm \omega}_{p})$.
In order to derive images $i_c({\bm \omega}_{p})$, we describe two existing MC approaches~\cite{marshak20053d}.
\begin{enumerate}
  \item Forward Monte Carlo (FMC): photons propagate from the source (sun) to the detector.
  \item Backward Monte Carlo (BMC): photons propagate from the detector to the source.
\end{enumerate}

\subsubsection{Sampling by Inverse Transform}
\label{sec:SamplinMC}

MC is stochastic. It treats scattering and extinction as random phenomena sampled from probability distributions. Random sampling at the heart of MC is realized by an {\em inverse transform}~\cite{devroye1986sample} of a specified probability density function.
We now briefly describe this mathematical process.
Let $u$ be a random number drawn from a uniform distribution in the unit interval: $u\sim{\cal U}[0,1]$. The number $u$ can be transformed into a random variable $\chi$, whose cumulative distribution function (CDF) is $F(\chi)$. The transform is defined by $\chi = F^{-1}(u)$, where $F^{-1}$ denotes the inverse of $F$. Specifically, consider a photon propagating in the atmosphere. The photon has high probability of propagating as long as $t$ is high, but
the probability diminishes as $t\rightarrow 0$. Thus Eq.~(\ref{eq:beer-lambert}) can be viewed as a probability density
function, whose CDF is
\begin{equation}
  \label{eq:Ftau}
  F(\tau)=\int_{0}^{\tau}\exp(-\tau')\derivsym{\tau'}=1-\exp(-\tau).
\end{equation}
Each photon then propagates to a random optical depth
\begin{equation}
  \label{eq:taurand}
  \tau^{\rm random}=F^{-1}(u)=-\ln(1-u).
\end{equation}

\subsubsection{Forward Monte Carlo Photon Tracking}
\label{sec:FMC}

In this approach, photons are generated at the source, illuminating the top of the atmosphere (TOA) uniformly.
Photons propagate from the TOA in direction ${\bm \omega_{\rm sun}}$. Each photon is traced through the atmosphere. Photons that happen to reach camera $c$ about direction ${\bm \omega}_{p}$ are counted as a contribution to $i_c({\bm \omega}_{p})$. A photon's life cycle is then defined by the following steps (Fig.~\ref{fig:mcgrid}):
\begin{figure*}[t!]
  \centering
  \yoavcomment{\includegraphics{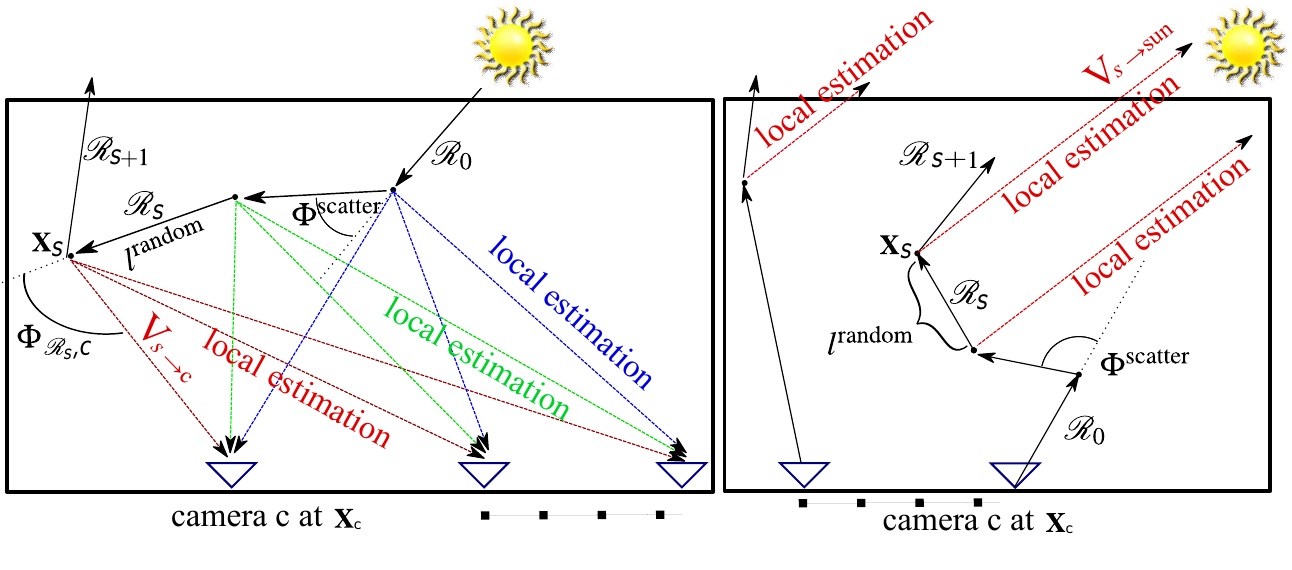}}
  \caption{\small [Left] Multi-view FMC with local estimation.
    [Right] Multi-view BMC with local estimation..}
  \label{fig:mcgrid}
\end{figure*}
~\noindent ({\tt i}) Launch a photon-packet from the TOA in
direction ${\bm \omega_{\rm sun}}$. This is the initial {\em ray}, denoted
${\cal R}_0$. The packet has initial intensity $I_0$.

\noindent Per iteration $s$:

~\noindent ({\tt ii}) Find the distance on ray ${\cal R}_{s}$ to which
the photon-packet propagates. Eq.~(\ref{eq:taurand})
yields $\tau^{\rm random}$. Then using Eq.~(\ref{eq:tau}), numerically seek
$l^{\rm random}$ s.t.
\begin{align}
  \int_0^{l^{\rm random}}(\sigma^{\rm aerosol} n +\beta^{\rm air}) dl
  = \tau^{\rm random}.
  \label{eq:MCextinct}
\end{align}
Distance $l^{\rm random}$ along ${\cal R}_{s}$ yields the 3D position ${\bf
  X}_{s}$.

~\noindent ({\tt iii}) If ${\bf X}_{s}$ is outside the domain, the
packet is terminated. If ${\cal R}_{s}$ passes through ${\bf X}_{c}$, or a small area around ${\bf X}_{c}$, the packet is counted as contributing to the image pixel.

~\noindent ({\tt iv}) Suppose ${\bf X}_{s}$ is inside the domain. The type of particle (air molecule or aerosol)
that the photon-packet interacts with at point ${\bf X}_{s}$ is randomly determined by the relative extinction coefficients ($\beta^{\rm air}$ vs. $\beta^{\rm aerosol}$) at the voxel containing ${\bf X}_{s}$.

~\noindent ({\tt v}) If the particle is an aerosol, the photon-packet intensity
is attenuated to $I_{s+1}=\varpi^{\rm aerosol}I_{s}$. For a purely scattering
particle, e.g. an air molecule, the photon-packet maintains its intensity. If $I_{s+1}$ is
lower than a threshold, the packet is stochastically terminated,
following~\cite{Iwabuchi2006}.

~\noindent ({\tt vi}) The photon-packet is scattered to a new random direction, determined by inverse transform sampling~\cite{frisvad2011importance,binzoni2006use}, according to the phase
function of the particle (Eqs.~\ref{eq:Phg},\ref{eq:Pray}). Let \mbox{$\Phi^{\rm scatter}={\rm arcos}({\bm \omega}\cdot {\bm \psi})$} be the off-axis scattering angle, relative to ${\bm \psi}$. Given a random sample $u\sim{\cal U}[0,1]$,
\begin{equation}
\label{eq:thetaHG}
\Phi^{\rm scatter}={\rm arcos}\bigg\{\frac{1}{2g}\Big[ 1+ g^{2}- \Big( \frac{g^{2}-1}{1+2 g  u-g }\Big) \Big]\bigg\}
\end{equation}
for an aerosol particles, and
\begin{align}
\label{eq:thetaREY}
\Phi^{\rm scatter}= &{\rm arcos}(\gamma^{\frac{1}{3}}-\gamma^{-\frac{1}{3}})
& &\textrm{with}& & \gamma=4u-2+\Big[  (4u-2)^{2} + 1 \Big]^{\frac{1}{2}}
\end{align}
for molecules. The scattering azimuth angle around ${\bm \psi}$ is sampled from ${\cal U}[0,2\pi]$. Following this scattering event, the photon traces a new {\em ray}, denoted ${\cal
  R}_{s+1}$, and the next iteration of propagation ({\tt ii}) proceeds.

\subsubsection{Local Estimation In FMC}
\label{sec:LE}

The quality of FMC increases with the number of photons launched.
Photons contributing to any pixel are accumulated in two ways. One way is step {\tt iii} above, which is a rare event. The second way is {\em Local estimation}~\cite{marshak20053d} which is used in conjunction to step {\tt vi}, in every scattering event (Fig.~\ref{fig:mcgrid}). The local estimation contributions $W_{\rm le}$ expresses the probability that a photon scatters towards the camera and reaches the camera without interacting again. Let ${\bf V}_{s\to \rm c}$ be the vector from the scattering point ${\bf X}_{s}$ to ${\bf X}_{c}$. Let $t_{s\to c}$ be the transmittance (\ref{eq:beer-lambert}) along ${\bf V}_{s\to c}$. Let $\Phi_{{\cal R}_{s},{\rm c}}$ be the angle between ${\cal R}_{s}$ and ${\bf V}_{s\to \rm c}$ (Fig.~\ref{fig:mcgrid}[left]). Let $\varpi$, $P$ be the respective albedo and phase function of the scattering particle. If a photon scatters by an aerosol, then $\varpi=\varpi^{\rm aerosol}$, $P=P_{\rm HG}$. If the photon scatters by a molecule then $\varpi=\varpi^{\rm air}\approx 1$, $P=P_{\rm Ray}$. Local estimation then contributes
\begin{align}
  W_{\rm le}=\varpi I_{s}P(\Phi_{{\cal R}_{s},{\rm c}})\frac{t_{s\to c}}{\left| {\bf V}_{s\to \rm c} \right|^{2}}
\label{eq:fle}
\end{align}
to pixel $p$ in camera $c$. The factor $\left| {\bf V}_{s\to \rm c} \right|^{-2}$ can be interpreted as consideration of ${\bf X}_{s}$ to be a point radiation source. Due to this factor, FMC is unstable when the camera is {\em in-situ} i.e, inside the scattering medium. Local estimation from scattering points ${\bf X}_{s}$ close to ${\bf X}_{c}$ lead to a large increase of image variance. Hence, an infinite number of photons is needed for convergence when $\left| {\bf V}_{s\to \rm c} \right| \to 0$.


\subsubsection{Backward Monte Carlo Photon Tracking}
\label{sec:BMC}

Numerically, BMC is very similar to the FMC. But, there are two major difference. First, from each pixel $p$ at camera $c$ a photon is separately launched in direction $-{\bm \omega}_{p}$. Then the photon is traced back through the atmosphere. A photon that happens to back-trace into the Sun, is counted as contribution to pixel $p$. The second difference is the local estimation calculation as we detail below.
BMC take the following steps (Fig.~\ref{fig:mcgrid}):

~\noindent ({\tt i}) Launch a photon-packet from camera $c$ to
direction $-{\bm \omega}_{p}$. This is the initial {\em ray}, denoted
${\cal R}_0$. The packet has an initial intensity $I_0$.
\noindent Per iteration $s$:

\noindent Per iteration $s$:

~\noindent ({\tt ii}) Find the distance on ray ${\cal R}_{s}$ to which
the photon-packet propagates, as described in step~({\tt ii}) of Sec.~\ref{sec:FMC}.

~\noindent ({\tt iii}) If ${\bf X}_{s}$ is outside the domain, the
packet is terminated. If ${\cal R}_{s}||{\bm \omega}_{\rm sun}$, the packet is counted as contributing to pixel $p$.

~\noindent ({\tt iv,v,vi}) Sample photons scattering event: particle type, photon-packet intensity, and scattered direction, as described in steps~({\tt iv,v,vi}) of Sec.~\ref{sec:FMC}.

\begin{figure*}[t!]
  \centering
  \yoavcomment{\includegraphics[scale=0.5]{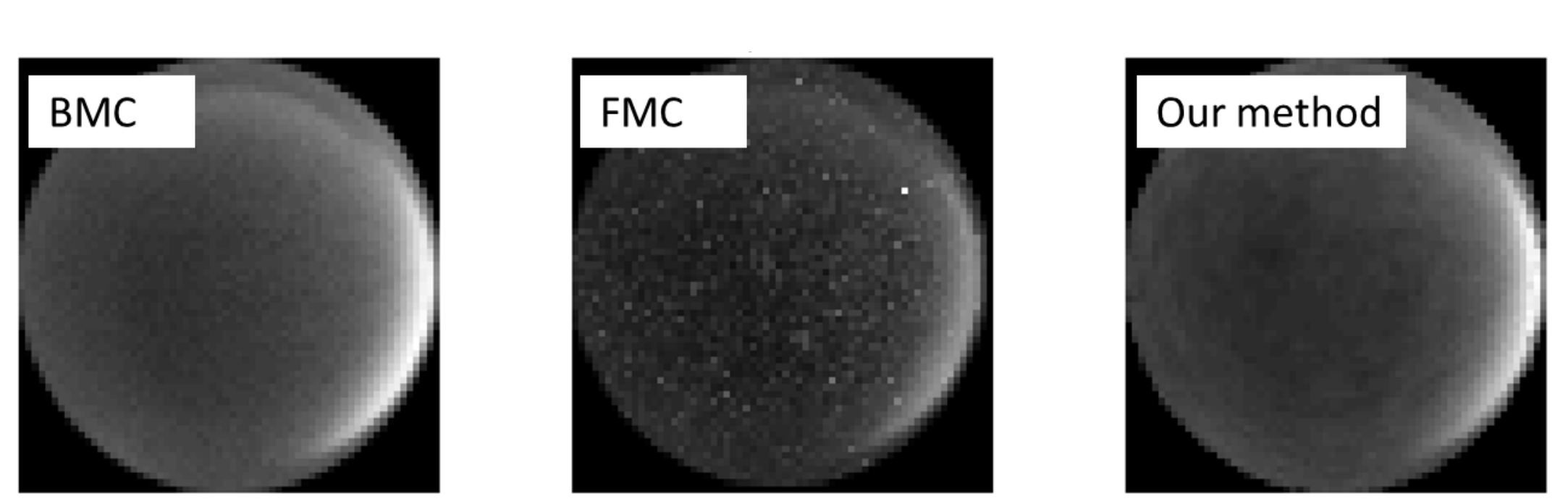}}
  \caption{\small Comparison of rendering results. In BMC, $10^{7}$ initial photons are used, equally divided bewteen all pixels that view the sky.
  In FMC, $10^{7}$ photons uniformly irradiate the domain TOA.}
  \label{fig:BMCcomparFMC}
\end{figure*}

Here too, local estimation is preformed in conjunction to step {\tt vi}. Here local estimation derives radiance back-traced to the sun, at each scattering event. Local estimation expresses the probability that a back propagating photon scatters towards the Sun, then reaches the Sun without interacting again.
Let ${\bf V}_{s\to \rm sun}$ be the vector from the scattering point ${\bf X}_{s}$ to the TOA, directed to $-{\bm \omega}_{\rm sun}$ (Fig.~\ref{fig:mcgrid}[right]). Here $t_{s\to \rm sun}$ is the transmittance along ${\bf V}_{s\to \rm sun}$, and $\Phi_{{\cal R}_{s},{\rm sun}}$ is the angle between ${\cal R}_{s}$ and ${\bf V}_{s\to \rm sun}$. Local estimation then contributes
\begin{equation}
  W_{\rm le}=\varpi I_{s}P(\Phi_{{\cal R}_{s},{\rm sun}})t_{s\to \rm sun}.
\label{eq:ble}
\end{equation}
Since the Sun is out of the scattering medium and effectively located at infinity, there is no $\left| {\bf V}_{s\to \rm sun} \right|^{-2}$ factor at all. Hence sky-images simulated by BMC are stable even in-situ. A comparison is displayed in Fig.~\ref{fig:BMCcomparFMC}: FMC rendering is very noisy compared to BMC.

\section{A Proposed Forward Model}
\label{sec:PFM}

As described in Sec.~\ref{sec:theor-backgr}, a BMC sky-image simulator has a major stability advantage over FMC. However BMC has drawbacks. BMC estimates radiance for one camera and one pixel at a time. In contrast, each single FMC sample trajectory can contribute to multiple viewpoints and pixels in parallel, using local-estimation (Fig.~\ref{fig:mcgrid}). This is efficient for simulating multiple cameras, which observe an atmospheric domain from $N_{\rm views}$ viewpoints.

We seek to use FMC for several reasons. First, FMC is more efficient for multi-pixel multi-view simulations. Moreover, a gradient-based recovery~\cite{gkioulekas2013inverse,levis2015},
which we need in Sec.~\ref{sec:IV} requires the volumetric fields $I({\bf X},{\bm \omega})$, $J({\bf X},{\bm \omega})$, not only projected images. Volumetric fields are obtained using FMC without local estimation, hence, are not prone to instabilities. To enable FMC in-situ, however, we need to overcome the $\left| {\bf V}_{s\to \rm c} \right|^{-2}$ instability. In this section, we describe a solution, disposing the $\left| {\bf V}_{s\to \rm c} \right|^{-2}$ factor using voxelization of the field $J$.
\begin{figure*}[t!]
  \centering
  \yoavcomment{\includegraphics[scale=0.18]{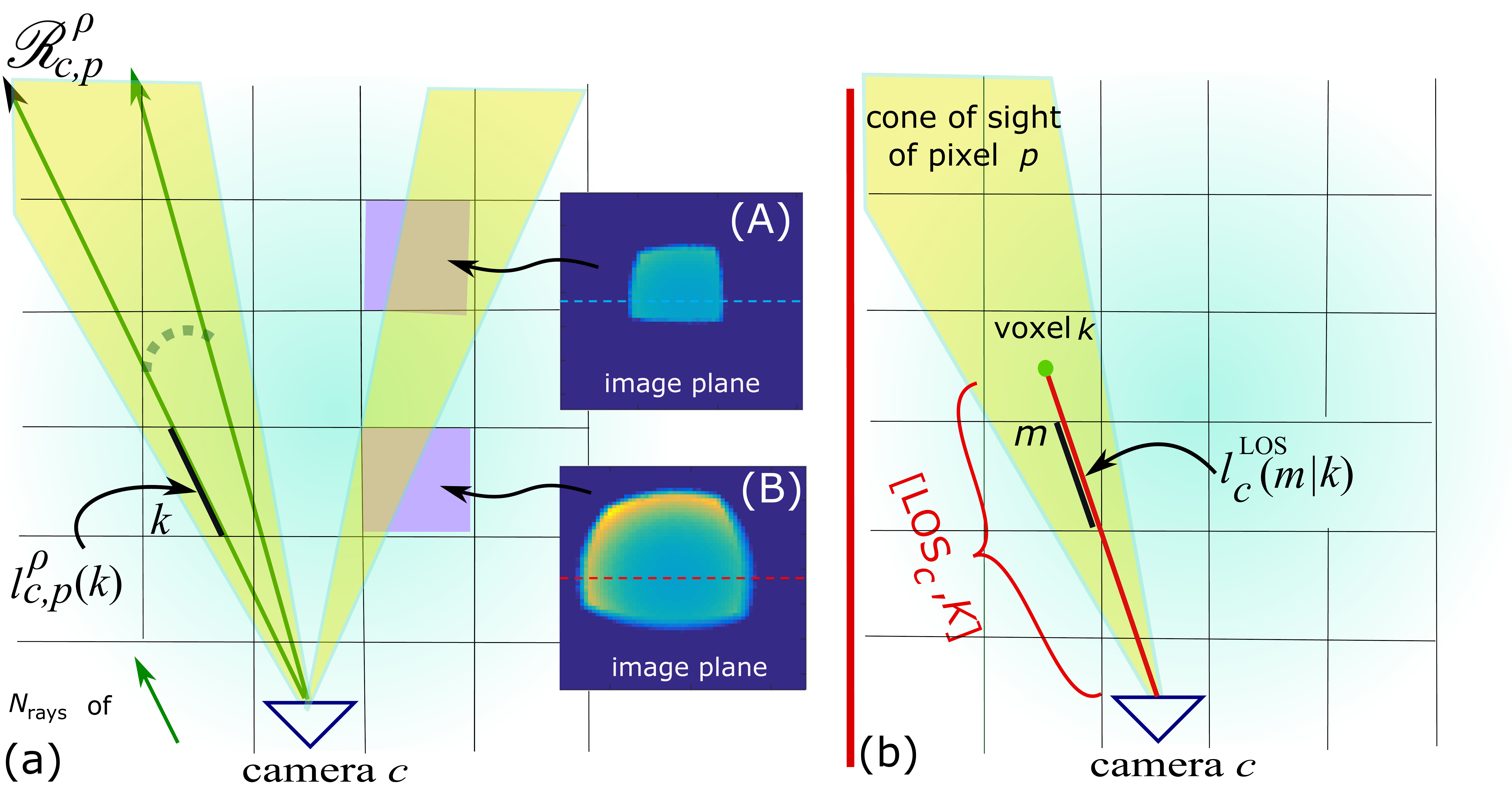}}
  \caption{\small (a) Ray ${\cal
  R}_{c,p}^{\rho}$ intersects with voxel $k$ creating a line-segment $l_{c,p}^{\rho}(k)$. Two isolated voxels (A) and (B) are projected to the image plane [Insets].
  Because of the difference of the voxels' distance from camera $c$, their projected support
  is different. (b) Line-segment $l_{c}^{\rm LOS}(m|k)$ is the intersection between voxel $m$ and line-segment $[{\rm LOS}_{c},k]$. }
  \label{fig:pixelrel}
\end{figure*}

As illustrated in Fig.~\ref{fig:setup} the volumetric domain is discretized into a grid of $N_{\rm voxels}$ rectangular cuboid voxels, indexed by $k$ or $m$. As a numerical approximation, assume that within any voxel, the parameters $\beta(k)$ ,${\sigma}^{\rm aerosol}$, $\varpi^{\rm aerosol}$, and $g$ are constants, e.g., corresponding to the values at each voxel center.

Our RT solution has three steps: ({\tt i})~Pre-calculate the geometry of the cameras-grid setup. ~({\tt ii})~FMC simulation calculates the radiance scattered from voxel $k$ in the direction of camera $c$. ~({\tt iii})~Light attenuation along the LOS from voxel $k$ to camera $c$.

\subsection{Geometry}
\label{sec:geo-data}
A camera sensor comprises of $N_{\rm pix}$ pixels. Each pixel collects light from a narrow cone in the atmosphere (Fig.~\ref{fig:pixelrel}a). The cone either contains or intersects some voxels, while remaining oblivious to the rest. The radiant power contributed by voxel $k$ to camera $c$ is $R_c(k)$, which we define in detail in Sec.~\ref{sec:optical-tran}. Overall, radiance captured at the pixel from all voxels is a weighted sum of $R_c(k)$ over all voxels $k$. This sum is
formulated by a sparse $N_{\rm pix}\times N_{\rm voxels}$ matrix operation ${\vect{\Pi}}_c$, having reciprocal area units
\begin{equation}
  {\bm i}_c= {\vect{\Pi}}_c{\bm R}_c
  \;\;.
  \label{eq:IcP}
\end{equation}
Here ${\bm i}_c$ is the image, column-stacked to a vector $N_{\rm
pix}$ long, and ${\bm R}_c$ is a column-stacked representation of $R_c(k)$. The weights of ${\vect{\Pi}}_c$ represent the relative portion of the radiant power contributing to camera $c$, solely due to geometry.

These weights are pre-calculated as follows: Divide pixel $p$ to $N_{\rm rays}$ points, from each of which back-project a ray ${\cal R}_{c,p}^{\rho}$. The intersection length of ${\cal R}_{c,p}^{\rho}$ (Fig.~\ref{fig:pixelrel}a) with voxel $k$ is $l_{c,p}^{\rho}(k)$. Let $V_{\rm voxel}$ be a voxel volume. The weight is then proportional to a normalized average intersection length,
\begin{align}
{\vect{\Pi}}_c(p,k) =&\bar{l}_{c,p}(k) = \frac{1}{N_{\rm rays}V_{\rm voxel}} \sum_{\rho=1}^{N_{\rm rays}} l_{c,p}^{\rho}(k)
\;\;.
\label{eq:PIc}
\end{align}
Eqs.~(\ref{eq:IcP},\ref{eq:PIc}) express rendering. There is no factor proportional to $\left|{\bf V}_{s\to \rm c}\right|^{-2}$ in Eqs.~(\ref{eq:IcP},\ref{eq:PIc}). This factor is implicit in the weighted sum matrix ${\vect{\Pi}}_c$: each voxel contributes to several pixels, illuminating a spot in the image plane. More rays pass through voxels closer to a camera. Thus, if a scattering event occurs in a voxel for which $\left|{\bf V}_{s\to \rm c}\right|$ is small, the contribution to the image affects more pixels than if the voxel had a large $\left|{\bf V}_{s\to \rm c}\right|$. This is expressed by a larger spot in the image  (Fig.~\ref{fig:pixelrel}a).

\subsection{Scattered Radiance Calculation with FMC}
\label{sec:scatter-radian}
Define the {\em scattered radiance} as $L({\bf X},{\bm \omega})=J({\bf X},{\bm \omega})\beta({\bf X})$. Using FMC, $L({\bf X},{\bm \omega})$ can be estimated by caching all the scattering events that occurred at ${\bf X}$ in direction ${\bm \omega}$. Our situation is simpler for two reasons. First, we use a voxelized radiance grid. Hence ${\bf X}$ is discretized to the voxel index $k$. Second, as shown below, we only need to store the scattered radiance that contributes to the discrete set of $N_{\rm views}$ cameras $c= 1,...,N_{\rm views}$. We denote the power scattered from voxel $k$ in the direction of camera $c$ by $L_c(k)$.
For each scattering event in voxel $k$, update $L_c(k)$ by
\begin{equation}
  L_c(k)\leftarrow L_c(k)+\varpi I_{s}P(\Phi_{{\cal R}_{s},{\rm c}}).
\label{eq:Lcupdate}
\end{equation}
Hence $L_c$ is discretized in space and the relevant directions.
Similarly, a discrete version of $J({\bf X},{\bm \omega})$ is
\begin{equation}
  j_c(k)=\frac{L_c(k)}{\beta(k)}.
\label{eq:Jc}
\end{equation}

\subsection{Optical Transmittance}
\label{sec:optical-tran}
The transmittance between ${\bf X}$ and ${\bf X}_{c}$ is
\begin{equation}
    t({\bf X},{\bf X}_{c})={\rm exp}\Big[-\int_{{\bf X}_{c}}^{{\bf X}}\beta(r)dr\Big].
\label{eq:tc}
\end{equation}
Eq.~(\ref{eq:IRTE}) can be re-written as image rendering:
\begin{equation}
\label{eq:IRTEL}
I({\bf X}_{c},{\bm \omega}_{p})=A+\int_{{\bf X}_{c}}^{{\bf X}_{\partial\Omega}}L({\bf X}',{\bm \omega}_{p})t({\bf X}',{\bf X}_{c})d{\bf X}',
\end{equation}
where $A$ represents direct solar rays entering the camera. For each camera $c$, denote by $[{\rm LOS}_c,k]$ a ${\rm LOS}$ between camera $c$ and the center of voxel $k$ (Fig.~\ref{fig:pixelrel}b). Suppose this LOS intersects voxel $m$. The geometric length of this intersecting line segment is $l^{\rm LOS}_{c}(m|k)$.
Following Eq.~(\ref{eq:tau}), the optical depth between the center of voxel $k$ to camera $c$ is
\begin{align}
  \tau_{{\rm LOS}_c}(k)&= \sum_{m \in[{\rm LOS}_{c},k]} l^{\rm LOS}_{c}(m|k)
  \beta(m).
  \label{eq:taoKtoc}
\end{align}

Define a $N_{\rm voxels}\times N_{\rm voxels}$ sparse matrix whose element $(k,m)$ is

\begin{equation}
  \OpDistance_{c}(k,m) =
  \left\{
    \begin{array}{ll}
      l^{\rm LOS}_{c}(m|k) & \mbox{ if $m \in[{\rm LOS}_c,k]$} \\
      0  & \mbox{ otherwise}
      \label{eq:Dvc}
    \end{array}
  \right.
  .
  \hspace{-0.05cm}
\end{equation}
Let $\vect{\tau}_{{\rm LOS}_c}$ and $\vect{\beta}$ be column-stack vector representations of $\tau_{{\rm LOS}_c}(k)$ and $\beta(k)$, respectively.  Then, we can write Eq.~(\ref{eq:taoKtoc}) using matrix notation
\begin{align}
  \vect{\tau}_{{\rm LOS}_c} &=
  \OpDistance_{c}\vect{\beta}.
\label{eq:bTck}
\end{align}
The discrete transmittance from the center of voxel $k$ towards camera $c$ is
\begin{equation}
T_{c}(k)=\exp[-\tau_{{\rm LOS}_c}(k)].
\label{eq:Tck}
\end{equation}
Based on Eqs.~(\ref{eq:IcP},\ref{eq:Lcupdate},\ref{eq:Tck})
 \begin{equation}
 R_{c}(k)= L_{c}
(k)T_{c}(k)\;.
  \label{eq:Rck}
\end{equation}
Let $\vect{T}_{c}$, $\vect{L}_c$ be the column stack vector representations of ${\rm T}_{c}(k)$ and ${\rm L}_{c}(k)$ respectively. A column-stack vector of all voxel contributions to camera $c$ is described by
\begin{equation}
  {\bm R}_c= \vect{L}_c \odot \vect{T}_{c}\;.
  \label{eq:Rc}
\end{equation}
Here $\odot$ denotes the Hadamard (element-wise) product.

Let ${\bm j}_{c}$ be a column-stack vector representation of $j_{c}(k)$. Then from Eqs.~(\ref{eq:IcP},\ref{eq:Jc},\ref{eq:Rc}), excluding direct sun light, the image is
\begin{equation}
  {\bm i}_c({\bm \beta})= {\vect{\Pi}}_c({\bm j}_c \odot {\bm Y}_{c})={\vect{\Pi}}_c({\bm j}_c \odot {\bm \beta} \odot{\bm T}_c)
  \;\;,
  \label{eq:IcPJbetaT}
\end{equation}
where ${\bm Y}_{c}= {\bm \beta} \odot {\bm T}_{c}$.
The value of pixel $p$ in image $c$ is
\begin{equation}
 i_c(p)= \frac{1}{N_{\rm rays}V_{\rm voxel}} \sum_{\rho=1}^{N_{\rm rays}} \sum_{k \in {\cal R}_{c,p}^{\rho}} l_{c,p}^{\rho}(k)
  j_{c}(k)\beta(k)T_{c}(k).
  \label{eq:Icpixel}
\end{equation}
Eqs.~(\ref{eq:IcP},\ref{eq:IcPJbetaT},\ref{eq:Icpixel}) are a discrete version of Eq.~(\ref{eq:IRTEL}), excluding the direct solar irradiance of the camera.

\section{Rendering Simulations}
\label{sec:Simulations}

\begin{figure}[t!]
  \centering
  \yoavcomment{\includegraphics[scale=0.3]{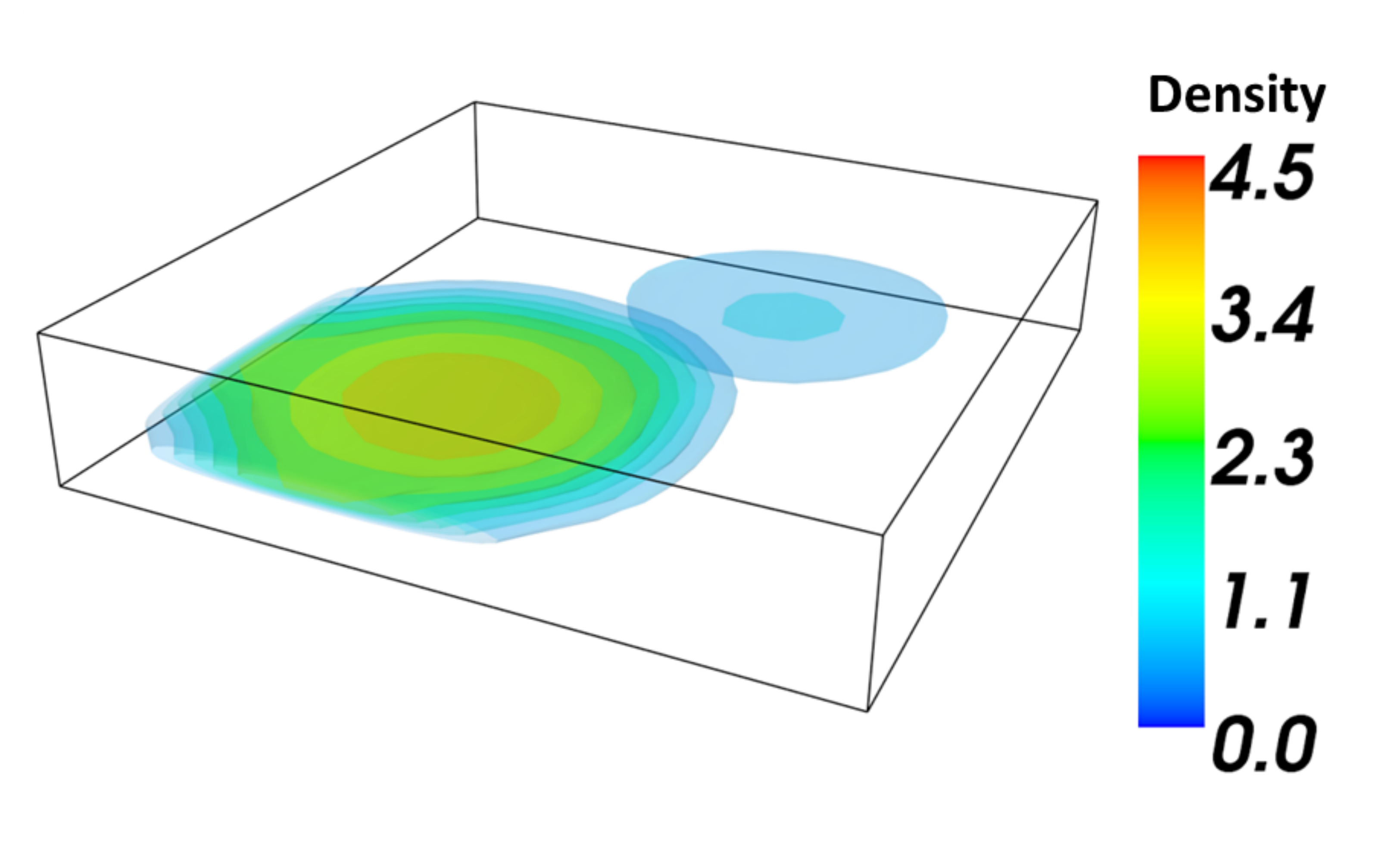}}
  \caption{\small Aerosol distributions~\cite{aides2013}. {\em Haze blobs} low density distribution.
  The aerosol density unit is $10^{6}~{\rm particles}/{\rm m}^3$.}
  \label{fig:gtdist}
\end{figure}

We tested the scenes used in~\cite{aides2013}, illustrated in Fig.~\ref{fig:gtdist}. We briefly re-mention them here for clarity.\\
\noindent {\bf Geometry}:
The atmospheric domain is $50{\rm km}^{2}\times50 {\rm km}^{2}$ wide, $10{\rm km}$ thick.  The fields $\beta^{\rm aerosol}({\bf X})$ and $\beta^{\rm air}({\bf X})$ from~\cite{aides2013} are discretized to a $20\times20\times 40$ voxel grid. For rendering using the our method, the domain is more finely divided into a $80\times80\times120$ voxel grid, where the dimensions of each voxel is $625\times 625\times 83{\rm m}$. The sun is at zenith angle $\Phi^{\rm SR}=45^o$. Following~\cite{aides2013}, the sun's red-green-blue wavelengths intensity ratios are $255:236:224$. There are $N_{\rm views}=36$ ground-based cameras placed uniformly with $\sim 7$km nearest-neighbor separation.\\
\noindent {\bf Aerosols}:
Two aerosols types were used all having $\varpi^{\rm aerosol}=1$:
  \begin{enumerate}
    \item An artificial aerosol having an isotropic phase function. The extinction cross sections are in the red-green-blue (RGB) channels $\sigma^{\rm
        aerosol}_{\rm R}=\sigma^{\rm aerosol}_{\rm G}=\sigma^{\rm aerosol}_{\rm B}=17$~\si[sticky-per]{\per\micro\meter\squared}.

    \item Type 6 from the aerosol list
         in~\cite{Martonchik2009}. The anisotropy parameter per color
        channel is $[g_{\rm R},g_{\rm G},g_{\rm B}]=[0.763,0.775,0.786]$. The extinction cross sections are $[\sigma^{\rm
        aerosol}_{\rm R}, \sigma^{\rm aerosol}_{\rm G}, \sigma^{\rm
        aerosol}_{\rm
        B}]=[16.5,16.2,15.9]$~\si[sticky-per]{\per\micro\meter\squared}.
  \end{enumerate}
Let $n^{\rm sealevel}$ be a density of aerosols at sea level. Here we give a short description of the atmospheres. More details are found in~\cite{aides2013}.
We simulated different aerosol distributions:
\begin{description}
 \item[\textbf{Atm1}] Haze
      blobs (Fig.~\ref{fig:gtdist}) of an isotropic aerosol, at low density ($n^{\rm sealevel}\approx 10^6$).
 \item[\textbf{Atm2}] Haze
      blobs of an anisotropic aerosol, at low density ($n^{\rm sealevel}\approx 10^6$).
 \item[\textbf{Atm3}] Haze
      blobs of an anisotropic aerosol, at high density ($n^{\rm sealevel}\approx 10^7$).
\end{description}
\begin{figure*}[t!]
  \centering
  \yoavcomment{\includegraphics[scale=0.34]{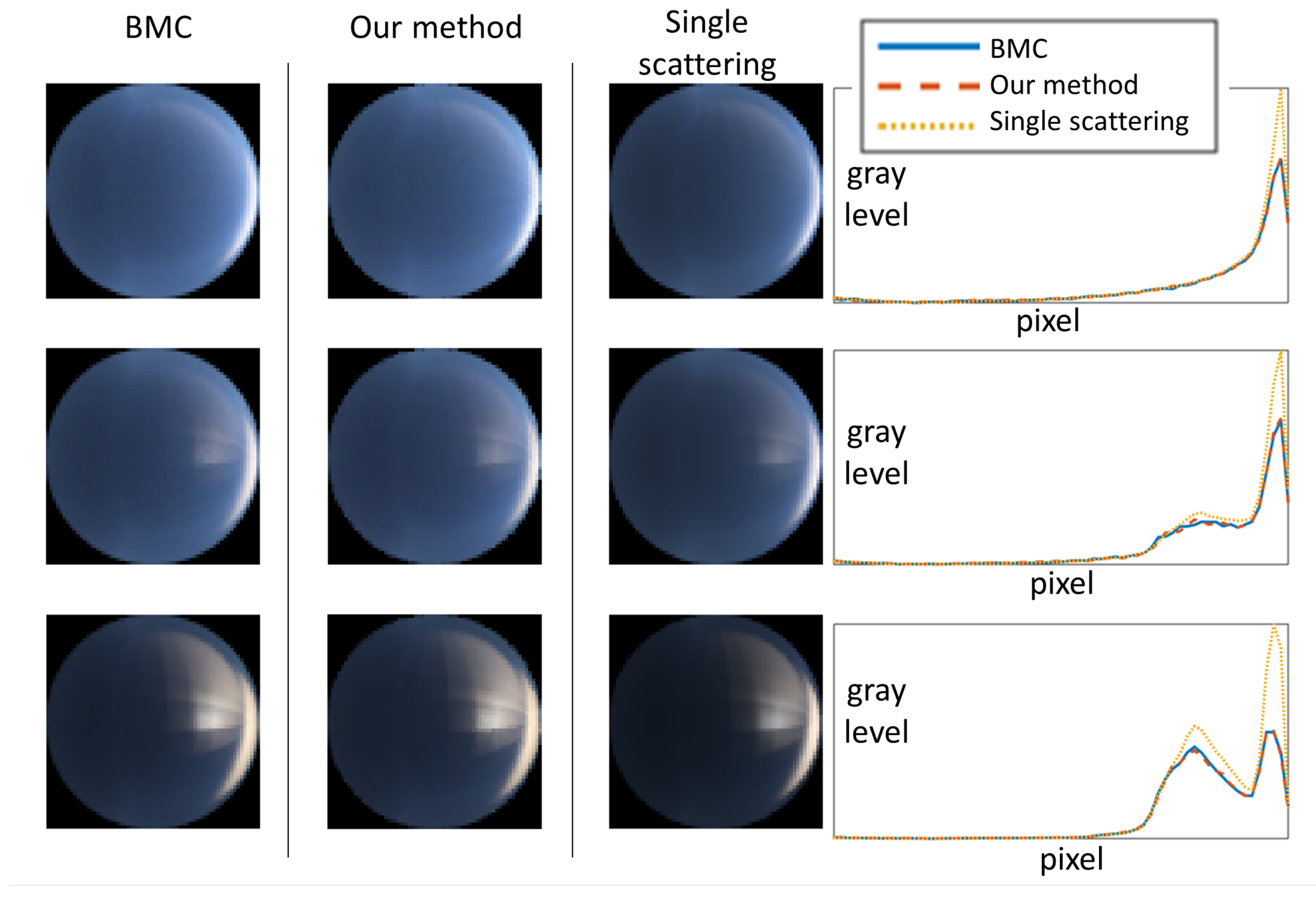}}
  \caption{\small Image rendering and a middle horizontal line cross-section (green channel).
[Top] Atm1, [middle] Atm2, and [bottom] Atm3.}
  \label{fig:mc3Renderings}
\end{figure*}
All the $N_{\rm views}$ imaging systems have a hemispherical field with $N_{\rm pix}=64\times64$.

Fig.~\ref{fig:BMCcomparFMC} demonstrates the improvement achieved by our method relative to simple FMC (Sec.~\ref{sec:FMC}), on a red channel image of Atm1.
Fig.~\ref{fig:mc3Renderings} compares images rendered using:
\begin{enumerate}
  \item BMC, initial $10^4$ photons per pixel, i.e, $\approx 1.5\cdot 10^9$ photons in total.
  \item Proposed voxelized FMC, using initial $10^7$ photons on the TOA, and $N_{\rm rays}=10$.
  \item Single scattering approximation~\cite{aides2013}.
\end{enumerate}
Rendering using our method is highly consistent with BMC rendering (Sec.~\ref{sec:BMC}), and similar to the single-scattering results~\cite{aides2013}. Deviations of MC from the single-scattering results are more pronounced where multi-scatter is more significant: near the horizon and generally in Atm3. There is normalization constant between the BMC and proposed method rendered images gray level. This constant doesn't depend on atmospheric matter parameters, but on the domain geometry, and was found empirically.

Processing was preformed using MATLAB on a 2.50~GHz Intel Xeon CPU. Rendering was parallelized with 40 cores.
Theoretically, For a given pre-calculated geometry (Sec.~\ref{sec:geo-data}) the proposed rendering method should be $\approx1.5\cdot10^9 / 10^7=150$ times faster. However, messaging between cores consumes additional run time, thus our method was 108 times faster than BMC.

\section{Inverse Problem}
\label{sec:IV}

The previous sections rendered images, assuming the field $\beta({\bf X})$ is known. Now we derive and solve the inverse-problem: given acquired images, what is $\beta({\bf X})$ ?
In addition to air molecules, let there be a single type of aerosol in the atmospheric domain. Hence, the three-element vector $[{\sigma}^{\rm aerosol},\varpi^{\rm aerosol},g]$ is uniform across the scene and assumed known~\cite{Martonchik2009}. The aerosol density is spatially variable and unknown, and so is $\beta^{\rm aerosol}({\bf X})\in \DistSet$.  Here $\DistSet$ comprises all possible extinction fields that comply with some constraints.
Particularly, ${\bm \beta}^{\rm aerosol}$ is non-negative
and its spatial support is bounded between the ground and the TOA.
A constraint useful for reducing the dimensionality is that ${\bm \beta}^{\rm aerosol}$ is
piecewise-constant, following 3D blocks of $N_{\rm x}\times N_{\rm y}\times N_{\rm z}$ voxels.
This constraint is consistent with an assumption that spatial variations of $\beta({\bf X})$ are generally smooth.

The data are $N_{\rm views}$ image measurements
 $\{ {\bm i}^{\rm measured}_c\}_{c=1}^{N_{\rm views}}$. Recovery is formulated as
an optimization of a cost function, to fit the image-formation model to the data~\cite{aides2013}.
\begin{align}
  \label{eq:minobjective}
  \hat{\vect{{\bm \beta}}} =
  \argmin_{{\bm \beta}\in\DistSet} \mathbfcal{E}({\bm \beta})
\end{align}

\subsection{Gradient-based Optimization}
\label{subsec:RA}

Using red-green-blue channels, the RGB extinctions are [${\bm \beta}_{\rm R},{\bm \beta}_{\rm G},{\bm \beta}_{\rm B}$]. We solve Eq.~(\ref{eq:minobjective}) using a gradient-based method. Let the variable extinction be ${\bm \beta}_{\rm G}$. Let $\mu\in[\rm R,\rm G,\rm B]$ be a channel color. Define $\tilde{\sigma_{\mu}}={\sigma}^{\rm aerosol}_{\mu}/{\sigma}^{\rm aerosol}_{\rm G}$. From the known $\sigma^{\rm aerosol}$ and ${\bm \beta}^{\rm air}$, ${\bm \beta}_{\rm G}^{\rm aerosol}$ determine the overall extinction per channel using
\begin{equation}
  \label{eq:channelBeta}
  {\bm \beta}_{\mu} =
  {\bm \beta}^{\rm air}_{\mu}+\tilde{\sigma_{\mu}}{\bm \beta}^{\rm aerosol}_{\rm G}.
\end{equation}
Let ${\bm i}_{c,\mu}$ and ${\bm i}^{\rm measured}_{c,\mu}$ denote the modeled and measured image in channel $\mu$, respectively.
Then, the cost function of Eq.~(\ref{eq:minobjective}) is
\begin{equation}
  \label{eq:Cost}
  \mathbfcal{E}({\bm \beta})
  = \sum_{\mu=\rm R,\rm G,\rm B}\sum_{c=1}^{N_{\rm views}}
  \left\|
    \MaskSun_{c}[{\bm i}^{\rm measured}_{c,\mu} - {\bm i}_c({\bm \beta}_{\mu})]
  \right\|^2_2  + \eta \Psi({\bm \beta}^{\rm aerosol}_{\rm G}).
\end{equation}
Here $\Psi$ is a regularization term that expresses the spatial smoothness~\cite{aides2013} of ${\bm \beta}^{\rm aerosol}_{\rm G}$, while $\eta$ is a regularization weight. In Eq.~(\ref{eq:Cost}), $\MaskSun_{c}$ represents masking of pixels around the Sun. There are two reasons for this masking. Sky-images estimated using MC have high variance in pixels surrounding the Sun due to the stochastic nature of MC. The problem occurs as the phase function is sampled at forward scattering angles.\footnote{To reduce noise in forward scattering, some works~\cite{Iwabuchi2006,Mayer2011} approximate the phase function or use {\em importance sampling}.} A sun-mask avoids use of noisy modeled pixel. Moreover, a sun occluder is typically applied cameras, to block lens flare and saturation~\cite{veikhermanclouds}.

Let $\transpose{(\cdot)}$ denote transposition. Then, the gradient of Eq.~(\ref{eq:minobjective}) is

\begin{align}
  \label{eq:divCost1}
  \frac{\partial{\mathbfcal{E}}}{\partial{\bm \beta}}
  = 2 \sum_{\mu=\rm R,\rm G,\rm B}\tilde{\sigma_{\mu}}\sum_{c=1}^{N_{\rm views}}
  {\bm Q}_{c}\transpose{\Big\{ \MaskSun_{c}{\cal J}_{c}({\bm \beta}_{\mu})
   \Big\}}
  \MaskSun_{c}[{\bm i}_{c}(\bm \beta_{\mu})-{\bm i}^{\rm measured}_{c,\mu}]
   + \eta \frac{\partial}{\partial{{\bm \beta}^{\rm aerosol}_{\rm G}}}\Psi({\bm \beta}^{\rm aerosol}_{\rm G}).
\end{align}
Here the matrix ${\cal J}_{c}({\bm \beta}_{\mu})$ is the Jacobian of the
vector ${\bm i}_{c,\mu}$ with respect to ${\bm \beta}_{\mu}$, and ${\bm Q}_{c}$ is a diagonal weighting matrix which is detailed in Sec.~\ref{sec:cond-opt}. Element $(p,k)$ of ${\cal J}_{c}({\bm \beta}_{\mu})$ differentiates the intensity of pixel $p$ in
viewpoint $c$ with respect to the extinction at voxel $k$,
\begin{equation}
{\cal J}_{c}({\bm \beta}_{\mu})=
\begin{bmatrix}
    \partial i_{c,\mu}(1)/\partial{\beta_{\mu}(1)} & \dots  & \dots  &\dots  & \partial i_{c,\mu}(1)/\partial{\beta_{\mu}(N_{\rm voxels})} \\
    \partial i_{c,\mu}(2)/\partial{\beta_{\mu}(1)} & \dots  & \dots  &\dots  & \partial i_{c,\mu}(2)/\partial{\beta_{\mu}(N_{\rm voxels})} \\
    \vdots & \vdots &  \vdots & \vdots & \vdots \\
    \partial i_{c,\mu}(p)/\partial{\beta_{\mu}(1)} & \vdots  &  \partial i_{c,\mu}(p)/\partial{\beta_{\mu}(k)} & \dots  & \dots \\
    \vdots & \vdots & \ddots  & \ddots & \ddots \\
    \partial i_{c,\mu}(N_{\rm pix})/\partial{\beta_{\mu}(1)} & \dots  &  \dots  & \dots  & \partial i_{c,\mu}(N_{\rm pix})/\partial{\beta_{\mu}(N_{\rm voxels})}
\end{bmatrix}.
\label{eq:DirectJacob}
\end{equation}
Estimating Eq.~(\ref{eq:DirectJacob}) using Eq.~(\ref{eq:IcPJbetaT}), is very complex. The reason is that Eqs.~(\ref{eq:JEq},\ref{eq:IRTEL}) express a recursive interplay of the fields $I$, $J$, that are functions of $\bm \beta$ (see Eqs.~\ref{eq:JbetaRTE},\ref{eq:IRTE},\ref{eq:IRTEL}). It is complex to perform recurtion per each gradient component.

In~\cite{gkioulekas2013inverse,aides2013}, a similar forward model is applied while recovery is done by least squares minimization.  The forward model in~\cite{aides2013} assumes single scattering for image rendering. Under the single scattering assumption, the unidirectional Sun rays scatter at most once, thus ${\bm j}_{c}$ is easily calculated: Eq.~(\ref{eq:JEq}) degenerates to an integral over a $\delta$ function of orientation. Then, the Jacobian ${\cal J}_{c}({\bm \beta}_{\mu})$ has a closed-form expression. Using MC for image rendering, there is no close-form solution for the gradient (\ref{eq:divCost1}).
The forward model in~\cite{gkioulekas2013inverse} uses FMC to render images of a homogeneous medium. Minimization in~\cite{gkioulekas2013inverse} is solved by a stochastic gradient descent, where the unknown parameters are the spatially uniform $\{\beta$,$\varpi$,$P(\cdot)\}$. Gradient computation in~\cite{gkioulekas2013inverse} is intensive, with a set of three cascade FMC simulations per optimization iteration. Trying a similar formulation in a heterogenous medium would mean $\mathbfcal{O}(N_{\rm voxels})$ FMC renderings per iteration, since there are $N_{\rm voxels}$ degrees of freedom. This approach is computationaly very expensive on large grids.

\subsection{An Efficient Approach}
\label{subsec:SA}

Instead of a direct estimation of Eqs.~(\ref{eq:divCost1},\ref{eq:DirectJacob}) we optimize ${\bm \beta}$ using a \emph{surrogate function}~\cite{levis2015}. Denote by ${\bm j}_{c,\mu}$ a field ${\bm j}_{c}$ in channel $\mu$. Let ${\bm j}_{\rm all}=\{\{ {\bm j}_{c,\mu}\}_{c=1}^{N_{\rm views}}\}_{\mu=\rm R,\rm G,\rm B}$. For a fixed ${\bm j}_{\rm all}$, Eq.~(\ref{eq:minobjective}) is easily minimized as we explain. Gradient-based optimization is iterative. Define ${\bm \beta}^{(q)}$ as an estimation of ${\bm \beta}$ in the $q$'th iteration.  Based on ${\bm \beta}^{(q)}$, the field ${\bm j}_{\rm all}^{(q)}$ is computed using FMC (Sec.~\ref{sec:scatter-radian}). This step is {\em not} an inverse problem but forward-model rendering. Consequently, the computational complexity of this step does not increase with $N_{\rm voxels}$.

\begin{figure*}[t!]
\centering
  \yoavcomment{\includegraphics[scale=0.45]{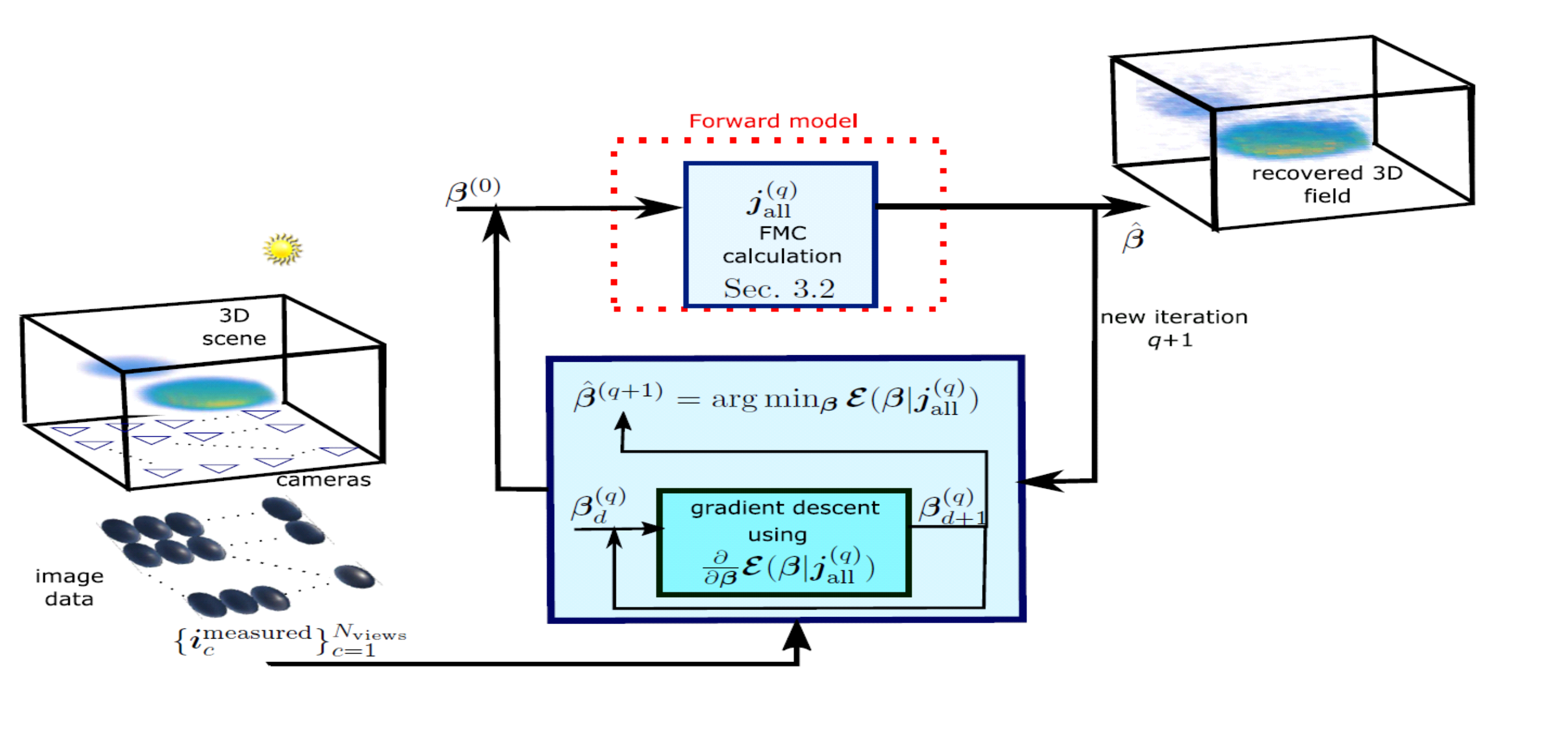}}
  \caption{\small Iterative optimization process using surrogate function $\mathbfcal{E}({\bm \beta}|{\bm j}_{\rm all}^{(q)})$.}
\label{fig:surrFun}
\end{figure*}

After ${\bm j}_{\rm all}^{(q)}$ is derived, it is fixed for a while. Denote $\mathbfcal{E}({\bm \beta}|{\bm j}_{\rm all}^{(q)})$ as a surrogate function for a fixed ${\bm j}_{\rm all}^{(q)}$. Keeping ${\bm j}_{\rm all}^{(q)}$ fixed, ${\bm \beta}$  is evolved. The following iterative optimization process is defined
\begin{align}
  \label{eq:minobjectiveBetaQplus1}
  \hat{{\bm \beta}}^{(q+1)} =
      \argmin_{{\bm \beta}\in\DistSet}
      \mathbfcal{E}({\bm \beta}|{\bm j}_{\rm all}^{(q)}).
\end{align}
Fig.~\ref{fig:surrFun} summarizes the iterative optimization process.

As we now show, when ${\bm j}_{\rm all}^{(q)}$ is fixed, the Jacobian (\ref{eq:DirectJacob}) degenerates to a simple calculation. We now detail the derivation of the Jacobian ${\cal J}_{c}({\bm \beta|{\bm j}^{(q)}_{c,\mu}})$ for a given fixed array ${\bm j}_{\rm all}^{(q)}$, i.e, $\partial{\bm j}^{(q)}_{c,\mu}/\partial{\bm \beta}_{\mu}\equiv 0$. Let ${\bm T}_{c,\mu}$ and ${\bm Y}_{c,\mu}$ be the fields ${\bm T}_{c}$ and ${\bm Y}_{c}$ in channel $\mu$, respectively.  Let $\OpDiag{\vect{v}}$ denote conversion of a general vector
$\vect{v}$ into a diagonal matrix, whose main diagonal elements
correspond to the elements of $\vect{v}$. Using Eqs.~(\ref{eq:bTck},\ref{eq:Tck},\ref{eq:IcPJbetaT}) and expressions from~\cite{aides2013} for the gradient of element-wise products, the Jacobian~(\ref{eq:DirectJacob}) degenerates to
\begin{align}
  \label{eq:Jaco1}
  {\cal J}_{c}({\bm \beta}_{\mu}|{\bm j}^{(q)}_{c,\mu})=\PartDeriv{{\vect{\Pi}}_c({\bm j}_{c,\mu}^{(q)} \odot {\bm Y}_{c,\mu})}{{\bm \beta}_{\mu}} =
  {\vect{\Pi}_c}\OpDiag{{\bm j}^{(q)}_{c,\mu}}\PartDeriv{{\bm Y}_{c,\mu}}{{\bm \beta}_{\mu}}
  ,
\end{align}
\begin{align}
  \label{eq:TcofJaco1}
  \PartDeriv{{\bm Y}_{c,\mu}}{{\bm \beta}_{\mu}} =
  \OpDiag{{\bm \beta}_{\mu}}\PartDeriv{{\bm T}_{c,\mu}}{{\bm \beta}_{\mu}}+
  \OpDiag{{\bm T}_{c,\mu}}
  ,
\end{align}
and
\begin{align}
  \label{eq:DivTc}
  \PartDeriv{{\bm T}_{c,\mu}}{{\bm \beta}_{\mu}} =
  - \OpDiag{\exp[-\OpDistance_{c}\vect{\beta}_{\mu}]}\OpDistance_{c}
  .
\end{align}

\begin{figure*}[t!]
\centering
  \yoavcomment{\includegraphics[scale=0.45]{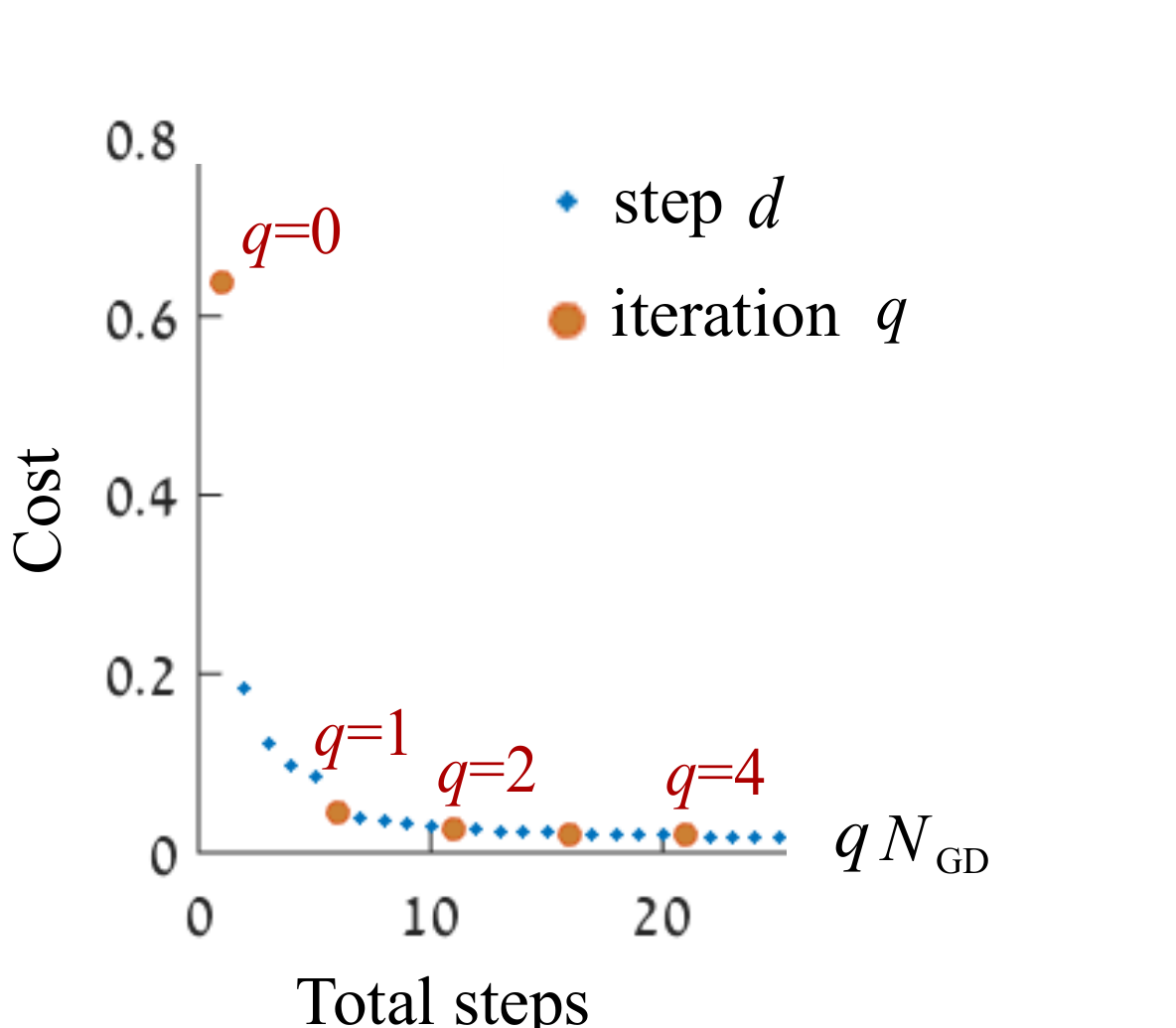}}
  \caption{\small Example of cost minimization using $N_{\rm GD}=5$.}
\label{fig:cost}
\end{figure*}

Given ${\cal J}_{c}({\bm \beta}_{\mu}|{\bm j}^{(q)}_{c,\mu})$, Eqs.~(\ref{eq:minobjectiveBetaQplus1},\ref{eq:divCost1}) can be solved using {\em Gradient Descent}
\begin{equation}
  \label{eq:GD}
  {\bm \beta}^{(q)}_{d+1} =
  {\bm \beta}^{(q)}_{d} -\Delta \frac{\partial}{\partial{{\bm \beta}}} \mathbfcal{E}({\bm \beta}|{\bm j}_{\rm all}^{(q)}),
\end{equation}
where $d$ indexes a gradient descent step, and $\Delta$ is the step size. After every $N_{\rm GD}$ gradient descent steps, ${\bm j}_{\rm all}$ is updated. Then, gradient-descent of $\bm \beta$ is resumed, using the updated ${\bm j}_{\rm all}$ for another set of $N_{\rm GD}$ gradient descents, and so on until convergence. Fig.~\ref{fig:cost} shows an example of the cost minimization using $N_{\rm GD}=5$.
A major advantage of the surrogate function is effective gradient calculation, without $N_{\rm voxels}$ rendering processes.

\section{Conditioning the Optimization}
\label{sec:cond-opt}

\begin{figure*}[t!]
  \centering
  \yoavcomment{\includegraphics[scale=0.30]{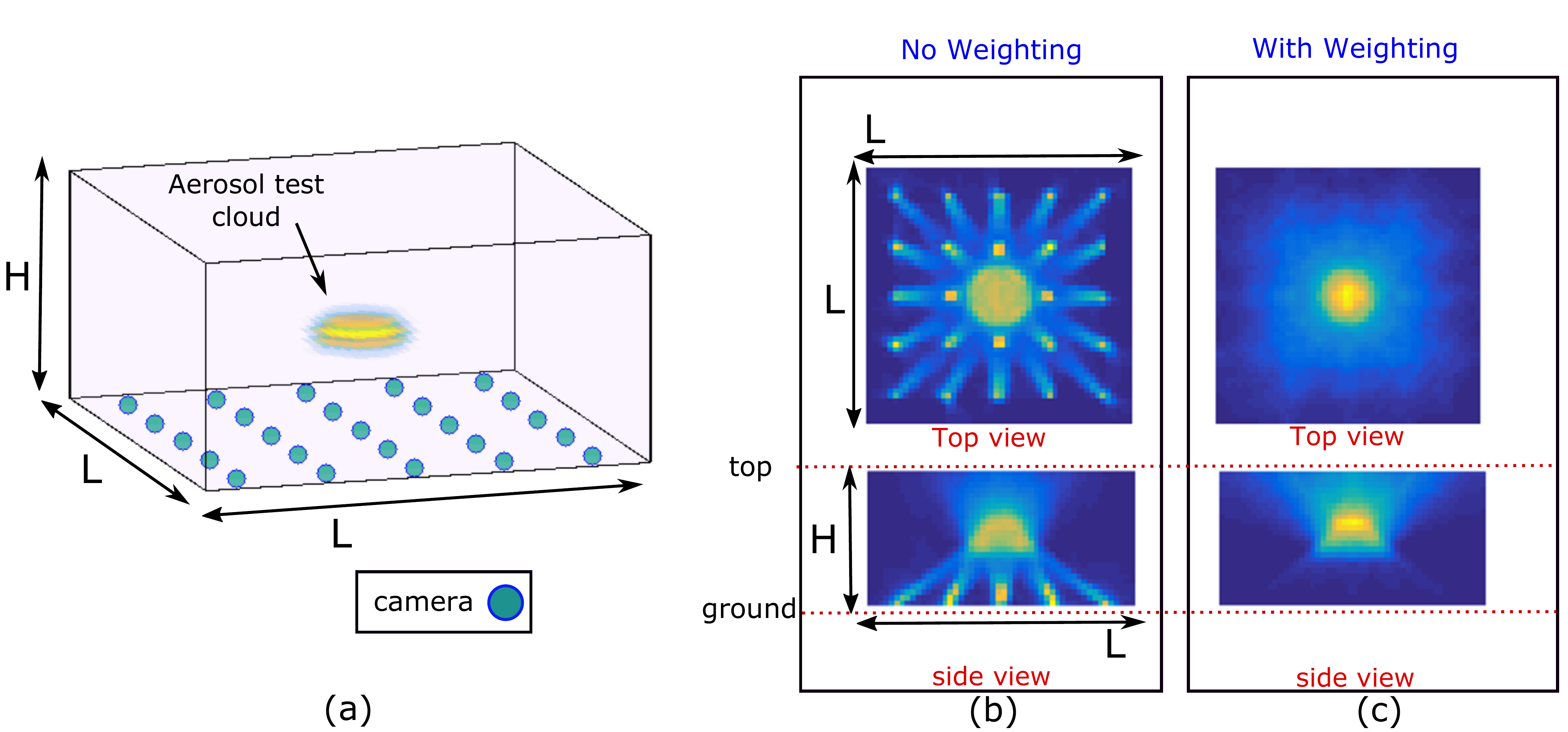}}
  \caption{\small (a) The domain is observed by 25 ground-based cameras. An elliptic aerosol cloud at the center of an air-filled domain. (b,c) Top view and side view MIP visualizations of  ${\bm \beta}_{1}^{(0)}$.}
\label{fig:gradshow}
\end{figure*}

Figure \ref{fig:gradshow}a illustrates a test atmosphere observed by 25 ground based cameras.
In Fig.~\ref{fig:gradshow}b, maximum intensity projections (MIP)~\cite{MAP1989} visualizes ${\bm \beta}_{1}^{(0)}$. This is the very first step $(d=1)$ in the first iteration $(q=0)$. The field ${\bm \beta}_{1}^{(0)}$ in Fig.~\ref{fig:gradshow}b stems from ${\bm j}_{\rm all}^{(0)}$, which was created by the initialization ${\bm \beta}^{\rm aerosol}=0$. The field contains artifacts. The artifacts appear as high values of ${\bm \beta}_{1}^{(0)}$ at voxels near the in-situ cameras. These voxels are unstable.
This problem is comparable to ill-conditioned linear optimization.

If all cameras are far from the scattering domain, all voxels are similary observed~\cite{levis2015}, and there is no conditioning problem. However, in-situ, voxels affect unequally the data (Fig.~\ref{fig:pixelrel}a). Due to geometry, if voxel $k$ projected to more pixels than voxel $m$ then, $\partial \mathbfcal{E} / \partial{{\beta}(k)}$ tend to be significantly higher than $\partial \mathbfcal{E} / \partial{{\beta}(m)}$. Depending on $\Delta$, this imbalance leads to instability in nearby voxels or very slow convergence.
This problem is solved by conditioning, achieved using a diagonal weighting matrix ${\bm Q}_{c}$. Element $(k,k)$ of  ${\bm Q}_{c}$ is the number of rays ${\cal R}_{c,p}^{\rho}$ that pass trough voxel $k$. The field ${\bm \beta}_{1}^{(0)}$ that evolves using this conditioning is visualized in Fig.~\ref{fig:gradshow}c. This field has much weaker artifacts.

In addition, Eq.~(\ref{eq:Cost}) includes a regularization term. Our regularization term~\cite{aides2013} is $\Psi({\bm \beta}^{\rm aerosol}_{\rm G}) = \| {\bf W} \Laplacian{{\bm \beta}^{\rm aerosol}_{\rm G}}\|^2_2$,
where $\Laplacian$ is a matrix representation of the 3D Laplacian operator. The matrix ${\bf W}$ is diagonal: its elements are a function of the altitude of each voxel~\cite{aides2013}. The gradient of $\Psi({\bm \beta}^{\rm aerosol}_{\rm G})$ which used in Eq.~(\ref{eq:divCost1}), is $2\transpose{\Laplacian}\transpose{{\bf W}}{\bf W}\Laplacian{\bm \beta}^{\rm aerosol}_{\rm G}$.


\section{Recovery Simulations}
\label{sec:inv-sim}

\subsection{Images and Noise}
\label{sec:images}

For the scenes described in Sec.~\ref{sec:Simulations}, a set $\{{\bm i}^{\rm measured}_c\}_{c=1}^{N_{\rm views}}$ was rendered using BMC (Sec.~\ref{sec:BMC}) since BMC is the most accurate and precise method, despite its slow speed. These images are noisy because MC sampling implicitly induces Poissonian noise. This naturally mimics photon noise in optical imaging. To fully simulate a camera, we incorporate scaling of optical energy to graylevels in a 10-bit camera, read noise and quantization. These operations are expressed by
\begin{equation}
  {\bm i}^{\rm measured}_c
   ~~  \leftarrow ~~
  {\bm i}^{\rm measured}_c
  \left(
   \frac{2^{10}}
       {{\rm max}
       \{ \{\MaskSun_{c} {\bm i}^{\rm measured}_c \}_{ c=1}^{N_{\rm views}}
       \}}
   \right)
   + o_{\rm noise}.
   \label{eq:noise}
\end{equation}
The read-noise by $o_{\rm noise}$ is white, with standard deviation of 0.4 graylevels.
The values in Eq.~(\ref{eq:noise}) are clipped to the range $[0\ldots 1024]$ and rounded.
The resulting images are the input for our reconstruction method and its comparison to~\cite{aides2013}.


\subsection{Recovery Results}
\label{sec:recov-results}

\begin{figure}[t!]
  \centering
  \yoavcomment{\includegraphics[width=\columnwidth]{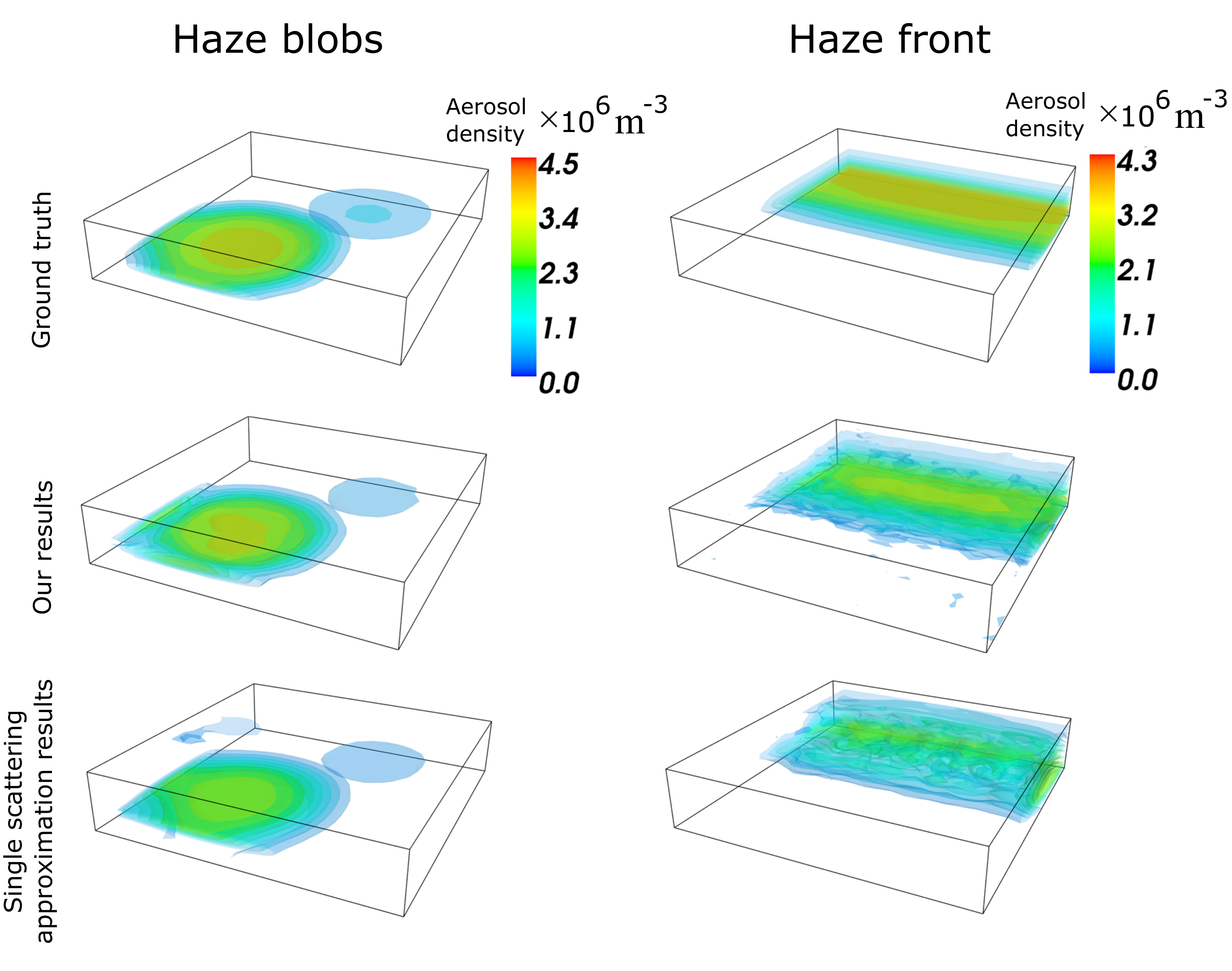}}
  \caption{\small [Top] Ground-truth aerosol distributions: Atm1 (Haze blobs, [left]), and Atm4 (Haze front, [right]). Color encodes aerosol density [$10^{6}~{\rm particles}/{\rm m}^3$]. [Middle] Our recontraction . [bottom] Reconstruction using the single scattering approximation~\cite{aides2013}.}
  \label{fig:distributions}
\end{figure}

In addition to the three aerosol distributions in Sec.~\ref{sec:Simulations}, the following distribution is used as in~\cite{aides2013}:
\begin{description}

 \item[\textbf{Atm4}] Haze Front of an anisotropic aerosol, at low density, shown in Fig.~\ref{fig:distributions} ($n^{\rm sealevel}\approx 10^6$).

\end{description}
The Haze Front is rendered as an elliptic cylinder intersected by the domain, as detailed in~\cite{aides2013}.

\begin{table}[t!]
  \centering
  \begin{tabular}{|c|cc|cc|cc|}
    \toprule
    Scene   & \multicolumn{2}{c|}{Single scattering}      &\multicolumn{2}{c|}{Our method}       & \multicolumn{2}{c|}{Our method, initialized} \\
            &\multicolumn{2}{c|}{}                 &\multicolumn{2}{c|}{null initialization}
    &\multicolumn{2}{c|}{by single scattering} \\[1ex]
            &$\delta_{\rm mass}$ &$\epsilon~~$ &$\delta_{\rm mass}$ &$\epsilon~~$ &$\delta_{\rm mass}$ &$\epsilon~~$\\

     \midrule
     Atm1   &1.7\%   &50\%  &3.4\%    &26\%   &3\%   &23\% \\
     Atm2   &-6.3\%  &61\%  &10\%     &38\%   &11\%   &37\% \\
     Atm3   &14\%    &63\%  &4.1\%    &27\%   &7\%   &28\%\\
     Atm4   &23\%    &76\%  &2.4\%   &70.8\%  &5.7\%   &43\% \\
    \bottomrule
  \end{tabular}
  \caption{Comparison of relative errors of our recovery method to single scattering approximation method from~\cite{aides2013}.}
  \label{tbl:mass}
\end{table}

The optimization was initialized by ${\bm \beta}^{\rm aerosol}=0$, or by a results~\cite{aides2013} obtained using single scattering approximation . The analysis used the following parameters: $\Delta = 10^{-3}$ and $N_{\rm GD}=5$. Starting from ${\bm \beta}^{\rm aerosol}=0$, satisfactory convergence occurred after several hundred iterations.

The total estimation error~\cite{aides2013} is quantified by the aerosol mass that is over and under-estimated
in all voxels, relative to the total aerosol mass in the scene. Using the $\ell_1$ norm,
the total mass relative difference is
$\delta_{\rm mass}=(\|\hat{\bm n}\|_1 - \|{\bm n}^{\rm true} \|_1) / \| {\bm n}^{\rm true} \|_1$.
In order to sense local errors~\cite{aides2013}, we use $\epsilon=\| \hat{\bm n} - {\bm n}^{\rm true} \|_1 / \| {\bm n}^{\rm true} \|_1$.
Some reconstructions are illustrated in Fig.~\ref{fig:distributions}. Additionally, Table~\ref{tbl:mass} summaries and compares the quantitative results for all the described distributions.
In conclusion, our results are better than~\cite{aides2013}. Thus, accounting for multiple scattering is important and significant, while being computationally feasible.
%

\noindent {
\textbf{Acknowledgements:}: We are grateful to Anthony Davis, Raanan Fattal,
Michael Zibulevsky for fruitful discussions.
We thank Mark Sheinin, Johanan
Erez, Ina Talmon, Dani Yagodin for support. YYS is a Landau
Fellow - supported by the Taub Foundation. His work
is conducted in the Ollendorff Minerva Center. Minerva is
funded through the BMBF. AL acknowledges ISF and ERC for financial support.
}

\end{document}